\documentclass{article} 
\usepackage{iclr2027_conference,times}
\usepackage{algorithm}
\usepackage{algpseudocode}
\usepackage{listings}  

\usepackage{amsmath,amsfonts,bm}

\def\eqref#1{equation~\ref{#1}}

\def\1{\bm{1}}

\DeclareMathAlphabet{\mathsfit}{\encodingdefault}{\sfdefault}{m}{sl}
\SetMathAlphabet{\mathsfit}{bold}{\encodingdefault}{\sfdefault}{bx}{n}

\usepackage{booktabs}   
\usepackage{multirow}   
\usepackage{colortbl}   
\usepackage{xcolor}     
\usepackage{adjustbox}  
\usepackage{enumitem}
\usepackage{hyperref}
\usepackage{url}
\usepackage{graphicx}

\newcommand{\cf}{\textit{cf}. }
\newcommand{\eg}{\textit{e}.\textit{g}.}
\newcommand{\method}{\textsc{Macr}}
\title{Multi-Agent Relay Thinking}

\author{
    Xu Liu\textsuperscript{\rm 1}\thanks{Equal contribution.},
    Jungang Qiao\textsuperscript{\rm 2}\footnotemark[1],
    Guikun Chen\textsuperscript{\rm 1},
    Xiaojuan Qi\textsuperscript{\rm 3},
    Wenguan Wang\textsuperscript{\rm 1}\thanks{Corresponding author.} \\
    \textsuperscript{\rm 1}Zhejiang University \\
    \textsuperscript{\rm 2}Nanyang Technological University \\
    \textsuperscript{\rm 3}The University of Hong Kong
}

\usepackage{wrapfig}
\usepackage{caption}

\iclrfinalcopy 
\begin{document}

\maketitle

\begin{abstract}

Existing multi-agent collaboration (MAC) methods have primarily explored a \textit{group thinking} collaboration topology, where agents independently generate solutions and aggregate their outputs. In this work, we investigate an alternative \textit{relay thinking} topology, where reasoning ownership is dynamically transferred across agents, allowing subsequent agents to inherit and refine an evolving reasoning trajectory. We introduce \textbf{M}ulti-\textbf{A}gent \textbf{C}onfidence-guided \textbf{R}elay (\method{}), a framework that enables dynamical relay collaboration among agents. The key challenge is determining when reasoning should be relayed and which agent should continue the process. \method{} addresses this challenge through confidence-guided relay triggering and a bidding mechanism that selects the next agent based on its confidence. 
Experiments on four benchmarks show that \method{} outperforms the strongest single-model strategies in accuracy, \eg, \textbf{+3.3\%} on MathVista and \textbf{+3.9\%} on MMStar, and remains on par with multi-agent strategies while consuming \textbf{6.0$\times$ less} token overhead on HallusionBench and MMStar.
Further analysis demonstrates that relay collaboration enables heterogeneous agents to progressively integrate complementary insights, combining heterogeneous capabilities across reasoning stages. Our source code is available at \href{https://github.com/67L1/MACR}{GitHub}.
\end{abstract}

\section{Introduction}
\label{sec:intro}
\begin{wrapfigure}[19]{r}{0.4\textwidth}
    \centering
    \small
    \vspace{-15pt}
    \includegraphics[width=\linewidth]{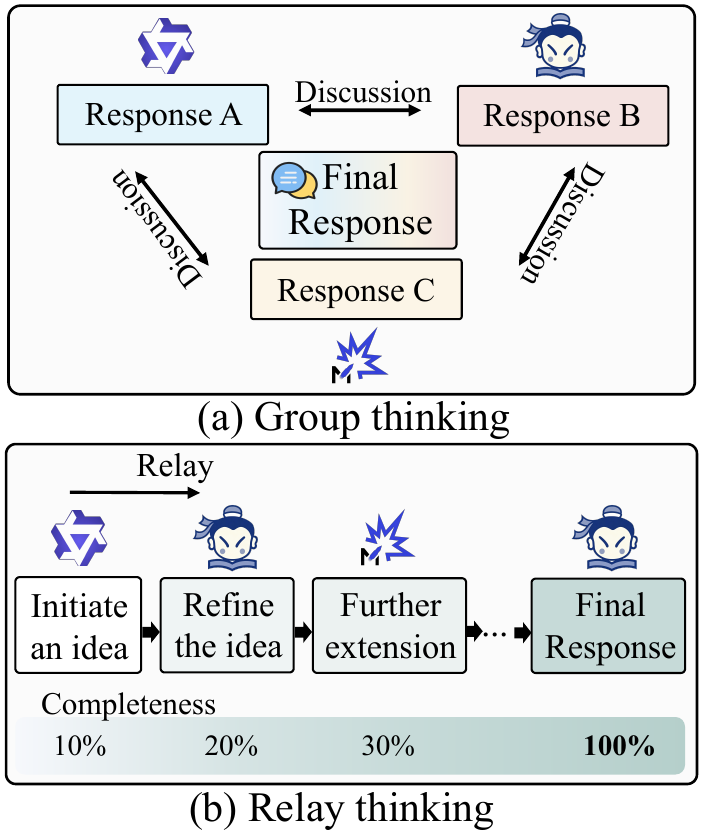}
    \captionsetup{width=0.4\textwidth}
        \vspace{-15pt}
    \caption{Comparison of group thinking\\ and relay thinking in MAC.}
    \label{fig:intro}
\end{wrapfigure}

\textbf{Background.} 
Large Language Models (LLMs) demonstrate impressive reasoning capabilities~\citep{zhang2025system,xu2025toward,liu2026let}. However, when solving complex problems, a single model might encounter intermediate reasoning bottlenecks that limit further progress~\citep{shojaee2026illusion,yan2026inftythink,liu2026sinktrack}. Multi-Agent Collaboration (MAC) has emerged as an effective approach to address such challenges by enabling multiple models to interact and exploit complementary capabilities~\citep{jin2025comprehensive,zhang2025cchall,bajoria2026language,tran2025multi,ferrag2026llm,abou2025agentic,gupta2025multi}. Existing studies have demonstrated the effectiveness of MAC in various domains, \eg, mathematical reasoning~\citep{qian2025modelingagent,davoudi2025collective}, visual question answering~\citep{wang2026towards,wang2026smr}, and multi-step decision making~\citep{weng2025benchform,nourzad2025dr}.

\textbf{Motivation.}
Despite their diversity, most MAC methods follow a \textit{group thinking} collaboration pattern~\citep{janis1982groupthink}, where multiple agents contribute reasoning trajectories in parallel and integrate information through interaction or aggregation~\citep{li2024improving,veluri2024beyond,becker2026stay} (\cf Fig.~\ref{fig:intro}a). This pattern enables agents to contribute parallel perspectives and reach a synthesized solution. However, collaboration among intelligent systems can take multiple forms. Cognitive science suggests another pattern of collective problem solving, referred to as \textit{relay thinking}~\citep{michael1998relay,van2025structuring}, where one participant develops an intermediate idea and transfers it to another participant for further refinement. Inspired by this sequential knowledge transfer process, we explore a relay-style collaboration topology for MAC, where agents progressively inherit and extend a shared reasoning trajectory through sequential ownership transitions (\cf Fig.~\ref{fig:intro}b).

\begin{figure}[t]
    \centering
    \includegraphics[width=0.95\linewidth]{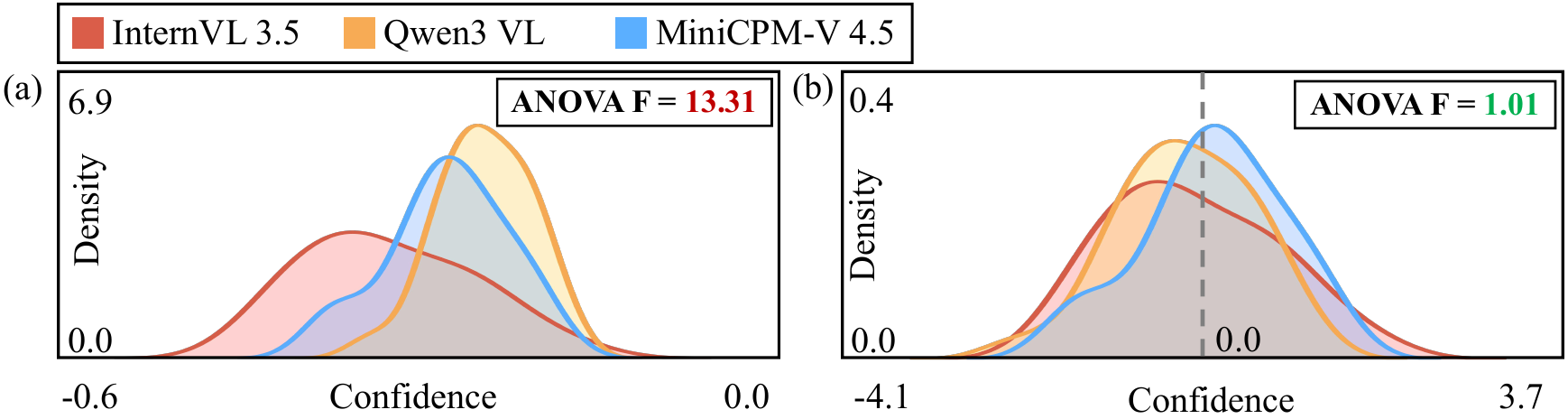}
    \caption{KDE of local confidence across three models, before (a) and after (b) calibration.}
    \label{fig:shift}
    \vspace{-15pt}  
\end{figure}
\begin{figure}[t]
    \centering
    \includegraphics[width=0.99\linewidth]{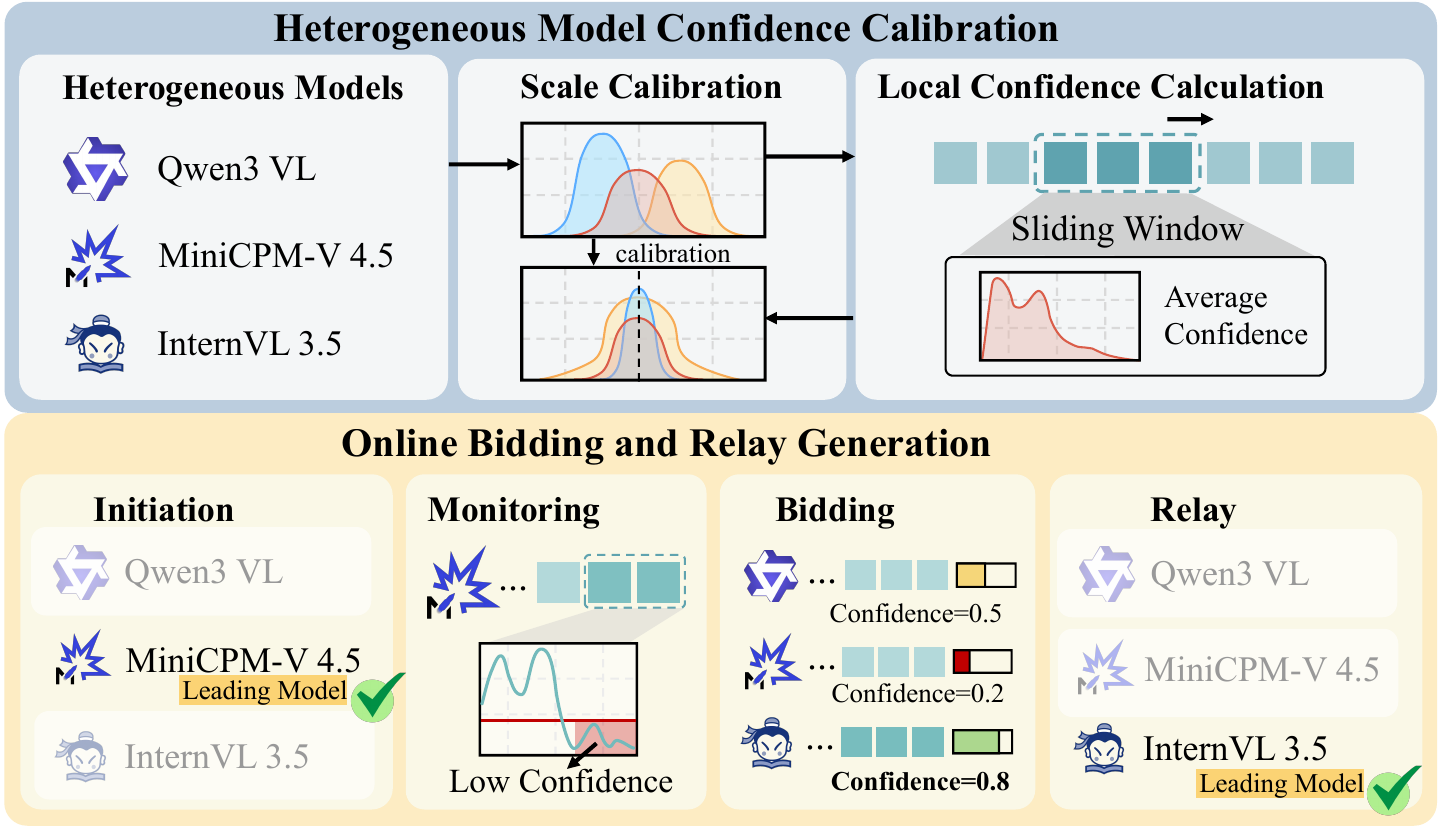}
    \put(-30.5,200.4){\footnotesize (Eq.~\ref{eq3})}
    \put(-101.5,215.4){\small \textbf{(\S\ref{sec:calibration})}}
    \put(-328,84.4){\footnotesize (Eq.~\ref{eq4})}
    \put(-228,84.4){\footnotesize (Eq.~\ref{eq5})}
    \put(-147,84.4){\footnotesize (Eq.~\ref{eq6})}
    \put(-54,84.4){\footnotesize (Eq.~\ref{eq7})}
    \put(-121,101.4){\small	 \textbf{(\S\ref{sec:bidding})}}

    \caption{An overview of \method{}'s two phases.}
    \label{fig:overview}
    \vspace{-10pt}  
\end{figure}

\textbf{Methodology.}
Realizing relay collaboration requires addressing two questions: when should reasoning ownership be transferred, and which agent should continue the reasoning process. We use model confidence as a signal for these decisions, motivated by previous studies showing that confidence estimates can provide information about reasoning reliability~\citep{xu2026unveiling,wang2026inference,farquhar2024detecting}. Moreover, we also observe in Fig.~\ref{fig:wc} that correct and incorrect trajectories are separable in confidence. However, confidence values derived from heterogeneous models are not directly comparable due to differences in model architectures, vocabulary spaces, and probability distributions (\cf Fig.~\ref{fig:shift}). To address this, we propose the \textbf{M}ulti-\textbf{A}gent \textbf{C}onfidence-guided \textbf{R}elay (\method{}). \method{} begins by extracting local confidence across a heterogeneous model pool using a sliding window for confidence calibration, reducing the Analysis of Variance (ANOVA) $F$ value from $13.31$ to $1.01$ (\cf Fig.~\ref{fig:shift}). On this basis of comparable confidence across heterogeneous models, \method{} monitors the leading model's calibrated confidence and interrupts generation to initiate bidding once that model shows low confidence, relaying generation rights to the model with the highest confidence. Besides deciding when and to whom to relay, \method{} must also determine which model should start the task. To this end, it initializes the leading model via an initial bidding round, selecting the model with the highest calibrated confidence to start the task. These steps together constitute \method{}'s online ``initiation–monitoring–bidding–relay'' loop (\cf Fig.~\ref{fig:overview}).

\textbf{Contributions.}
First, we offer a new perspective on MAC patterns by drawing on \textit{relay thinking}, where a reasoning trajectory is progressively refined and synthesized as it is relayed across heterogeneous models.
Second, we develop \method{}, a confidence-guided relay framework that calibrates confidence across heterogeneous models and jointly determines model initiation and reasoning handoffs through bidding.
Third, we reveal two empirical properties underlying effective heterogeneous collaboration: models exhibit complementary strengths across reasoning stages, while reasoning bottlenecks emerge at different stages across tasks. We hope these findings motivate dynamic model scheduling that adapts both reasoning ownership and relay timing throughout inference.

\begin{figure}[t]
    \centering
    \includegraphics[width=1.0\linewidth]{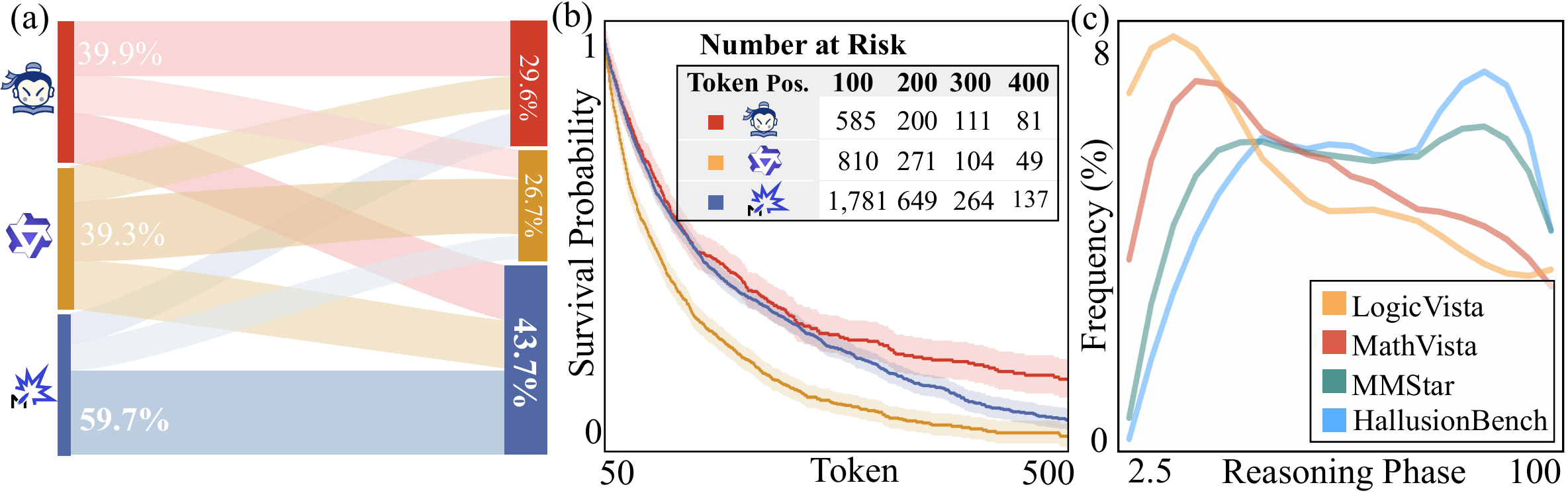}
    \caption{Analysis experiments of \method{}'s performance.}
    \label{fig:ana}
    \vspace{-5pt}
\end{figure}

\textbf{Results and Analysis.}
Experiments on four multimodal benchmarks show that \method{} outperforms single-model strategies in accuracy, e.g., \textbf{+3.3\%} on MathVista and \textbf{+3.9\%} on MMStar, and remains on par with multi-agent strategies while consuming \textbf{6.0$\times$ fewer} tokens on HallusionBench and MMStar, respectively. Qualitative analysis reveals that \method{} integrates the isolated cues of single models into a global reasoning chain(\cf \S\ref{sec:case}). Furthermore, our analyses confirm that heterogeneous models in \method{} are complementary, both across their capabilities (\cf Fig.~\ref{fig:ana}a) and across reasoning stages(\cf Fig.~\ref{fig:ana}b), and that they are capable of localizing task-agnostic bottlenecks(\cf Fig.~\ref{fig:ana}c).
\vspace{-5pt}  
\section{Methodology}
\vspace{-7pt}  
\label{sec:methodology}
\method{} requires determining two things. First, it requires the real-time identification of local reasoning bottlenecks. While a model's logarithmic probabilities indicate its confidence, we find that their confidence is incomparable. Therefore, models' confidence needs calibration to enable cross-model bidding. Second, it must evaluate which candidate model is best suited to take over generation. Candidate models perform short trial generations on a shared context, and \method{} evaluates their confidence under the calibrated metric to execute handovers.

Consequently, the implementation of \method{} is naturally divided into two sequential phases: \textbf{i}) Establish a well-calibrated confidence metric offline, ensuring reliable comparability among heterogeneous models (\S\ref{sec:calibration}); \textbf{ii}) Establish an online scheduling workflow comprising ``initiation–monitoring–bidding–relay'' stages to facilitate real-time collaboration (\S\ref{sec:bidding}).

\subsection{Heterogeneous Model Confidence Shift Analysis and Calibration}
\label{sec:calibration}

\noindent \textbf{Confidence Shift Analysis.} \method{} collaborates over a model pool $\mathcal{M}$ of three models (\textit{Qwen3 VL},\textit{ InternVL 3.5}, \textit{Minicpm-V 4.5}). We randomly sample 1,000 instances from MMMU~\citep{yue2024mmmu} and allow these models to perform reasoning. To obtain local confidence rather than sequence confidence for identifying reasoning bottlenecks, given a model $m \in \mathcal{M}$ and its generated logarithmic probability sequence $c_i$, we define a sliding window of size $w$ to calculate the local confidence:
\begin{equation}
    W_{i, m}^{(w)} = \frac{1}{w} \sum\nolimits_{j=0}^{w-1} c_{i-j}.
\label{eq1}
\end{equation}
We then extract three models' $W_{m}^{(w)}$ and visualize them via Kernel Density Estimation (KDE). The KDE visualizations show the distributions of the three models, where we observe clear shifts among them (\cf Fig.~\ref{fig:shift}a).
To formally verify this, we conduct a one-way analysis of variance (ANOVA). The $F$-value measures the ratio of between-group variance to within-group variance, quantifying whether a difference exists among the confidence distributions of the three models:
\begin{equation}
F = \frac{\frac{1}{k-1}\sum_{m=1}^{k} n_m(\bar{x}_m - \bar{x})^2}{\frac{1}{N-k}\sum_{m=1}^{k}\sum_{i=1}^{n_m}(x_{mi} - \bar{x}_m)^2},
\label{eq2}
\end{equation}
where $k$ is the number of candidate models, $n_m$ is the sample size for model $m$, $N$ is the total sample size, $\bar{x}_m$ is the confidence mean for the model $m$, and $\bar{x}$ is the overall mean. The experiment yielded $F = 13.31$ ($F \gg 1$), confirming that between-group differences significantly outweigh within-group random noise. This demonstrates that their raw confidence is mathematically incomparable.

\noindent \textbf{Confidence Calibration.} To eliminate the shift, a standardization method based on historical sliding window distributions is employed. Specifically, we aggregate the local confidence samples $W^{(w)}$ generated by each model on a historical reference dataset. This yields the local baseline mean $\mu_m^{(w)}$ and standard deviation $\sigma_m^{(w)}$ for window size $w$. During generation, each model's local confidence is transformed into a standardized metric:
\begin{equation}
    Z_{i,m}^{(w)} = \frac{W_{i, m}^{(w)} - \mu_m^{(w)}}{\sigma_m^{(w)}}.
\label{eq3}
\end{equation}
Subsequently, the ANOVA is conducted to verify the removal of shift (\cf Fig.~\ref{fig:shift}b). The KDE distribution redrawn with standardized data shows that the curves of the three models align around 0. Quantitatively, the F-value drops to $F = 1.01$ ($F \approx 1$),  indicating that between-group differences are reduced to the level of intra-model random noise, such that the shift across models is eliminated.
\begin{figure}[t]
    \centering
    \includegraphics[width=0.95\linewidth]{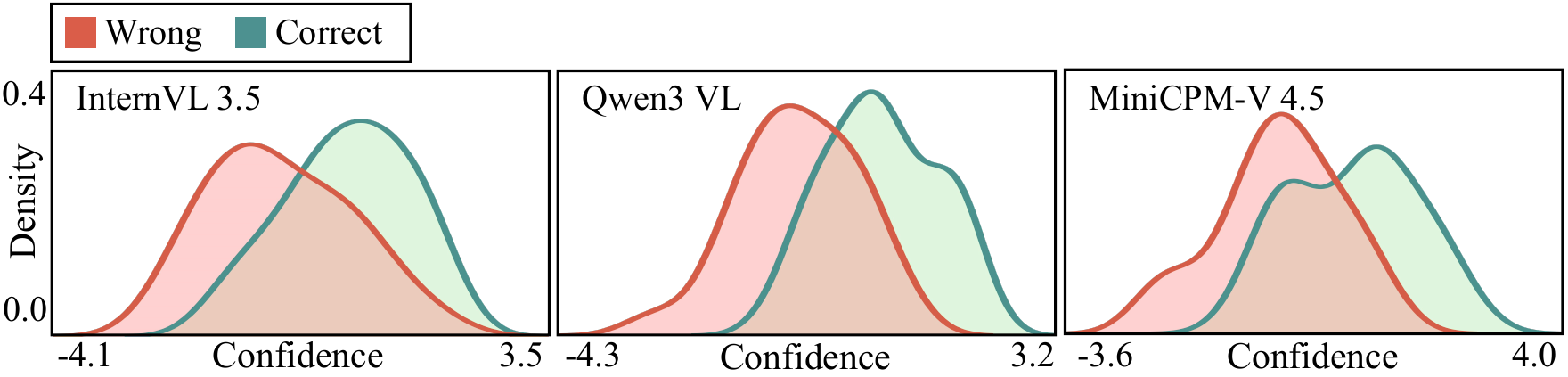}
    \caption{KDE of standardized local confidence grouped by correctness.}
    \label{fig:wc}
    \vspace{-15pt}  
\end{figure}
Furthermore, grouping the standardized data by correctness and visualizing their distributions via KDE (\cf Fig.~\ref{fig:wc}) reveals that correct trajectories exhibit higher local confidence. This separability is consistent with prior studies~\citep{xu2026unveiling,wang2026inference,farquhar2024detecting}, jointly confirming that a low confidence indicates a model encountering a bottleneck. Consequently, the historical expected mean and standard deviation for each model $m$ are adopted for cross-model calibration in the scheduling phase.

\subsection{Online Bidding and Relay Generation}
\label{sec:bidding}
\method{} applies the calibrated confidence metric (\S\ref{sec:calibration}) to identify local bottlenecks during models' generation, executing the ``initiation–monitoring–bidding–relay'' loop mechanism.

\noindent \textbf{Initiation.} To establish the initial leading model, all candidate models $m \in \mathcal{M}$ generate a length-$w$ sequence $T_m = \{t_{m,1}, t_{m,2}, \dots, t_{m,w}\}$ based on the initial context $X_0$. The mean local logarithmic probability for each sequence is calculated:
\begin{equation}
W_m^{(w)} = \frac{1}{w} \sum\nolimits_{j=1}^{w} \log p_m(t_{m,j} \vert X_0, t_{m,<j}),
\label{eq4}
\end{equation}
and calibrated to $Z_m^{(w)}$ using the model's $\mu_m^{(w)}$ and $\sigma_m^{(w)}$. \method{} assigns the initial autoregressive generation dominance to the model $m$ with that maximizes $Z_m^{(w)}$.

\noindent \textbf{Monitoring.} The leading model $m^*$ then performs standard autoregressive generation. At each time-step $t$, \method{} monitors the local confidence and calibrates it to $Z_{t, m^*}^{(w)}$:
\vspace{-3pt}
\begin{equation}
    \quad Z_{t, m^*}^{(w)} = \frac{\frac{1}{w} \sum_{j=0}^{w-1} c_{t-j} - \mu_{m^*}^{(w)}}{\sigma_{m^*}^{(w)}}.
\label{eq5}
\end{equation}
The sliding window calculation enables the identification of reasoning bottlenecks in real-time. Building on the separability of correct trajectories (\cf Fig.~\ref{fig:wc}), we define a threshold $\tau$. If the leading model's confidence drops below this threshold ($Z_{t, m^*}^{(w)} < \tau$), \method{} determines  it encounters a reasoning bottleneck, halting its generation to trigger the bidding.

\noindent \textbf{Bidding.} \method{} synchronizes the current context $X_t$ to each model $m \in \mathcal{M}$, prompting it to generate a $w$-token candidate sequence $T_m = \{t_{m,1}, t_{m,2}, \dots, t_{m,w}\}$. Evaluating their confidence, \method{} selects the model $m^*$ that maximizes the calibrated confidence:
\begin{equation}
    m^* = \arg\max_{m \in \mathcal{M}} 
    \big( \frac{\frac{1}{w} \sum_{j=1}^{w} \log p_m(t_{m,j} \mid X_t, t_{m,<j}) - \mu_m^{(w)}}{\sigma_m^{(w)}} \big).
\label{eq6}
\end{equation}
\noindent \textbf{Relay.} After selecting the model with the highest confidence, \method{} directly concatenates its sequence $T_{m^*}$ to the existing context, with each model independently maintaining its own KV cache so that no repeated prefilling is needed:
\begin{equation}
X_{t+w} = X_t \oplus T_{m^*}.
\label{eq7}
\end{equation}
\method{} then relays generation ownership to $m^*$ and resumes monitoring (\cf Eq.~\ref{eq5}) until completion. The algorithm flow is presented in Alg.~\ref{alg:macr_framework_main}.
\begin{algorithm}[t]
\renewcommand\thealgorithm{1}
\caption{Algorithm flow of \method{} framework.}
\label{alg:macr_framework_main}
\begin{algorithmic}[1]
\Require $q$ (query), $\mathcal{M}=\{m_1,\dots,m_N\}$ (model pool), $\tau$ (threshold), $w$ (window size)

\Function{MACR}{$q,\mathcal{M},\tau,w$}
    \State $c \gets q$ \Comment{context}
    \State $Z \gets \{\text{Calib}(m,c) \mid m \in \mathcal{M}\}$ \Comment{initial confidences}
    \State $m^* \gets \text{Best}(\mathcal{M},Z)$ \Comment{initiation}

    \While{\textbf{not} $\text{Done}(c)$}
        \State $x,z \gets \text{Monitor}(m^*,c,w)$ \Comment{token $x$, window confidence $z$}

        \If{$z<\tau$}
            \State $B \gets \{(X_m,z_m) \mid (X_m,z_m) = \text{Bid}(m,c),\, m \in \mathcal{M}\}$ \Comment{bid set}
            \State $m^*,X^* \gets \text{Best}(\mathcal{M},B)$ \Comment{relay}
            \State $c \gets \text{App}(c,X^*)$ \Comment{append relayed token sequence $X^*$}
        \Else
            \State $c \gets \text{App}(c,x)$ \Comment{append token}
        \EndIf
    \EndWhile

    \State \Return $c$
\EndFunction
\end{algorithmic}
\end{algorithm}
\section{Experiment}
\label{sec:exp}
\vspace{-8pt}
\subsection{Experimental Setup}
\label{sec:expsetting}
\vspace{-6.5pt}
\textbf{Datasets.\!} To evaluate \method{} across diverse dimensions, we adopt four benchmarks. \textbf{MMStar}~\citep{chen2024we} evaluates comprehensive multi-modal capabilities across 6 abilities and 18 tasks. \textbf{HallusionBench}~\citep{guan2024hallusionbench} diagnoses the visual and knowledge hallucinations. \textbf{LogicVista}~\citep{xiao2024logicvista} assesses structured reasoning in visual contexts, covering inductive, deductive, numerical, spatial, and causal logic. \textbf{MathVista}~\citep{lu2024mathvista} evaluates joint visual-mathematical reasoning across geometry, charts, and word problems.
\vspace{-2pt}

\textbf{Baselines.} We compare \method{} against single-model and multi-agent strategies. The model pool consists of \textit{Qwen3-VL-8B-Instruct}~\citep{yang2025qwen3}, \textit{InternVL3.5-8B}~\citep{wang2025internvl3}, and \textit{MiniCPM-V-4.5-8B}~\citep{yu2026minicpm}.

\vspace{-2pt}
$\bullet$ \textbf{Single-Model Strategies.} We evaluate standard input–output prompting (\texttt{IO}), where the model directly outputs the solution and answer from the given images and question; \texttt{CCoT}~\citep{mitra2024compositional} prompts the LLM to generate a scene graph capturing objects, attributes, and their relationships first, and then uses that scene graph to produce the final response; and \texttt{DDCoT}~\citep{zheng2023ddcot} decomposes complex problems into actionable sub-tasks and then produces the final answer.

\vspace{-2pt}
$\bullet$ \textbf{Multi-Agent Strategies.} We benchmark against two multi-agent strategies: Majority Vote (\texttt{MV})~\citep{wang2023selfconsistency}, establishing consensus across independent model generations, and \texttt{H-DMAD}~\citep{liu2025breaking}, a diverse multi-agent collaboration framework that assigns different reasoning strategies (\texttt{CoT}~\citep{zheng2023ddcot}, \texttt{CCoT}, and \texttt{DDCoT}) to different agents, encouraging complementary perspectives to break the fixed mental set, and fuses their outputs into a final answer.

\vspace{-2pt}
\textbf{Implementation Details.} We set the sliding window size $w=16$ and the confidence threshold $\tau=-0.3$ via parameter search on 600 randomly sampled instances from RealWorldQA~\citep{realworldqa}, which are disjoint from all evaluation sets. All evaluations are conducted under identical settings (\eg, temperature). To ensure reliability and rule out the influence of generation randomness, we conduct all evaluations with three distinct random seeds. We report accuracy as the mean $\pm$ standard deviation, and additionally report token usage along with the Incremental Benefit per Cost (IBC), defined as $\Delta\text{Acc} / \Delta\text{Tokens}$ relative to each single model's \texttt{IO} baseline~\citep{aggarwal2024automix}.

\begin{table*}[t]
\setlength{\tabcolsep}{4pt}
\centering
\renewcommand{\arraystretch}{1.25}
\begin{adjustbox}{width=0.99\textwidth}
\begin{tabular}{l ccc ccc ccc ccc}
\toprule
\multirow{2}{*}{\textbf{Model}}
& \multicolumn{3}{c}{\textbf{MathVista (1,000)}}
& \multicolumn{3}{c}{\textbf{LogicVista (448)}}
& \multicolumn{3}{c}{\textbf{MMStar (1,500)}}
& \multicolumn{3}{c}{\textbf{HallusionBench (951)}} \\
\cmidrule(lr){2-4} \cmidrule(lr){5-7} \cmidrule(lr){8-10} \cmidrule(lr){11-13}
& Acc & Tokens & IBC & Acc & Tokens & IBC & Acc & Tokens & IBC & Acc & Tokens & IBC \\
\midrule

\rowcolor{yellow!5!orange!5}
\multicolumn{13}{c}{\rule{0pt}{2.4ex}\textbf{\textsc{Single-Model Strategies}}} \\
\midrule

\rowcolor{black!2}
\multicolumn{13}{c}{\rule{0pt}{2.2ex}{\textit{Qwen3 VL~\citep{yang2025qwen3}}}} \\
\midrule
\texttt{IO}
& 70.50\tiny${\textcolor{gray}{\pm0.10}}$ & 1,117 & -
& 51.34\tiny${\textcolor{gray}{\pm0.22}}$ & 1,343 & -
& 65.93\tiny${\textcolor{gray}{\pm0.07}}$ & 901 & -
& 72.66\tiny${\textcolor{gray}{\pm0.11}}$ & 1,132 & - \\
\texttt{CCoT}
& 71.00\tiny${\textcolor{gray}{\pm0.17}}$ & 2,164 & 4.78
& 52.23\tiny${\textcolor{gray}{\pm0.39}}$ & 2,350 & 8.84
& 66.60\tiny${\textcolor{gray}{\pm0.12}}$ & 1,771 & 7.70
& 72.87\tiny${\textcolor{gray}{\pm0.18}}$ & 2,780 & 1.27 \\
\texttt{DDCoT}
& 71.70\tiny${\textcolor{gray}{\pm0.20}}$ & 2,671 & 7.72
& 53.13\tiny${\textcolor{gray}{\pm0.22}}$ & 3,188 & 9.70
& 67.60\tiny${\textcolor{gray}{\pm0.13}}$ & 2,135 & 13.53
& 73.61\tiny${\textcolor{gray}{\pm0.21}}$ & 2,634 & 6.32 \\

\midrule
\rowcolor{black!2}
\multicolumn{13}{c}{\rule{0pt}{2.2ex}{\textit{InternVL 3.5~\citep{wang2025internvl3}}}} \\
\midrule
\texttt{IO}
& 73.80\tiny${\textcolor{gray}{\pm0.26}}$ & 2,382 & -
& 51.79\tiny${\textcolor{gray}{\pm0.39}}$ & 2,392 & -
& 67.20\tiny${\textcolor{gray}{\pm0.18}}$ & 2,588 & -
& 71.82\tiny${\textcolor{gray}{\pm0.28}}$ & 2,262 & - \\
\texttt{CCoT}
& 74.20\tiny${\textcolor{gray}{\pm0.30}}$ & 5,210 & 1.41
& 52.46\tiny${\textcolor{gray}{\pm0.22}}$ & 5,206 & 2.38
& 68.00\tiny${\textcolor{gray}{\pm0.20}}$ & 5,434 & 2.81
& 71.92\tiny${\textcolor{gray}{\pm0.11}}$ & 4,913 & 0.38 \\
\texttt{DDCoT}
& 74.00\tiny${\textcolor{gray}{\pm0.10}}$ & 4,928 & 0.79
& 53.35\tiny${\textcolor{gray}{\pm0.39}}$ & 4,837 & 6.38
& 67.67\tiny${\textcolor{gray}{\pm0.07}}$ & 5,227 & 1.78
& 72.77\tiny${\textcolor{gray}{\pm0.18}}$ & 4,706 & 3.89 \\

\midrule
\rowcolor{black!2}
\multicolumn{13}{c}{\rule{0pt}{2.2ex}{\textit{MiniCPM-V 4.5~\citep{yu2026minicpm}}}} \\
\midrule
\texttt{IO}
& 74.40\tiny${\textcolor{gray}{\pm0.17}}$ & 607 & -
& 53.79\tiny${\textcolor{gray}{\pm0.22}}$ & 756 & -
& 66.67\tiny${\textcolor{gray}{\pm0.12}}$ & 510 & -
& 73.19\tiny${\textcolor{gray}{\pm0.21}}$ & 586 & - \\
\texttt{CCoT}
& 74.70\tiny${\textcolor{gray}{\pm0.20}}$ & 1,736 & 2.66
& 54.24\tiny${\textcolor{gray}{\pm0.39}}$ & 2,155 & 3.22
& 68.13\tiny${\textcolor{gray}{\pm0.13}}$ & 1,564 & 13.85
& 73.29\tiny${\textcolor{gray}{\pm0.28}}$ & 1,912 & 0.75 \\
\texttt{DDCoT}
& 74.50\tiny${\textcolor{gray}{\pm0.26}}$ & 1,396 & 1.27
& 54.02\tiny${\textcolor{gray}{\pm0.22}}$ & 1,801 & 2.20
& 67.33\tiny${\textcolor{gray}{\pm0.18}}$ & 1,269 & 8.70
& 73.61\tiny${\textcolor{gray}{\pm0.11}}$ & 1,424 & 5.01 \\

\midrule

\rowcolor{yellow!5!orange!5}
\multicolumn{13}{c}{\rule{0pt}{2.4ex}\textbf{\textsc{Multi-Agent Strategies}}} \\
\midrule

\texttt{MV}
& 77.80\tiny${\textcolor{gray}{\pm0.30}}$ & 7,699 & 7.80
& 55.13\tiny${\textcolor{gray}{\pm0.39}}$ & 7,943 & 4.54
& 71.00\tiny${\textcolor{gray}{\pm0.20}}$ & 7,508 & 7.19
& 74.13\tiny${\textcolor{gray}{\pm0.18}}$ & 8,072 & 2.45 \\

\texttt{H-DMAD}
& 78.80\tiny${\textcolor{gray}{\pm0.10}}$ & 23,780 & 2.63
& 55.58\tiny${\textcolor{gray}{\pm0.22}}$ & 27,574 & 1.27
& 73.20\tiny${\textcolor{gray}{\pm0.07}}$ & 24,677 & 2.83
& 75.50\tiny${\textcolor{gray}{\pm0.21}}$ & 26,375 & 1.19 \\

\rowcolor{teal!10}
\textbf{\method{}}
& 78.00\tiny${\textcolor{gray}{\pm0.17}}$ & 4,811 & \cellcolor{teal!25}\textbf{15.38}
& 55.80\tiny${\textcolor{gray}{\pm0.39}}$ & 6,149 & \cellcolor{teal!25}\textbf{7.89}
& 72.00\tiny${\textcolor{gray}{\pm0.12}}$ & 4,150 & \cellcolor{teal!25}\textbf{21.35}
& 74.66\tiny${\textcolor{gray}{\pm0.28}}$ & 4,414 & \cellcolor{teal!25}\textbf{7.71} \\

\bottomrule
\end{tabular}
\end{adjustbox}
\caption{Performance of MLLMs in terms of accuracy (\%), token overhead and IBC ($\times 10^{-6}$).}
\label{tab:mainres}
\vspace{-15pt}
\end{table*}
\vspace{-3pt}
\subsection{Quantitative Comparison Results}
\label{sec:mainres}
\vspace{-5pt}  
The results demonstrate that \method{} consistently surpasses single-model strategies in accuracy while remaining comparable to multi-agent strategies with minimal token overhead (\cf Table~\ref{tab:mainres}). \method{} yields substantial accuracy gains over the strongest single-model strategies, boosting performance by \textbf{+3.30\%} on MathVista (78.00\% vs. 74.70\%) and \textbf{+3.87\%} on MMStar (72.00\% vs. 68.13\%). It secures these enhancements while maintaining commensurate or reduced token overhead. For instance, \method{} demands only 4,150 tokens on MMStar, while single-model strategies like \texttt{CCoT} and \texttt{DDCoT} consume 5,434 and 5,227 tokens, respectively.
Furthermore, while multi-agent strategies achieve comparable accuracy, their token overhead is far higher. \method{} operates with \textbf{6.00 $\times$ fewer} tokens on HallusionBench (4,414 vs. 26,375) and MMStar (4,150 vs. 24,677). To quantify the efficiency gain of \method{}, we evaluate the IBC~\citep{aggarwal2024automix}. \method{} dominates IBC across all datasets with a state-of-the-art IBC of 21.35, \eg, boosting performance by +18.52 on MMStar (21.35 vs. 2.83) and +12.75 on MathVista (15.38 vs. 2.63) over \texttt{H-DMAD}, substantiating that \textit{relay thinking} successfully maximizes collective intelligence while preventing generation waste.

\vspace{-6pt}
\subsection{Qualitative Analysis}
\label{sec:case}
\begin{figure}[t]
    \centering
    \includegraphics[width=1.0\linewidth]{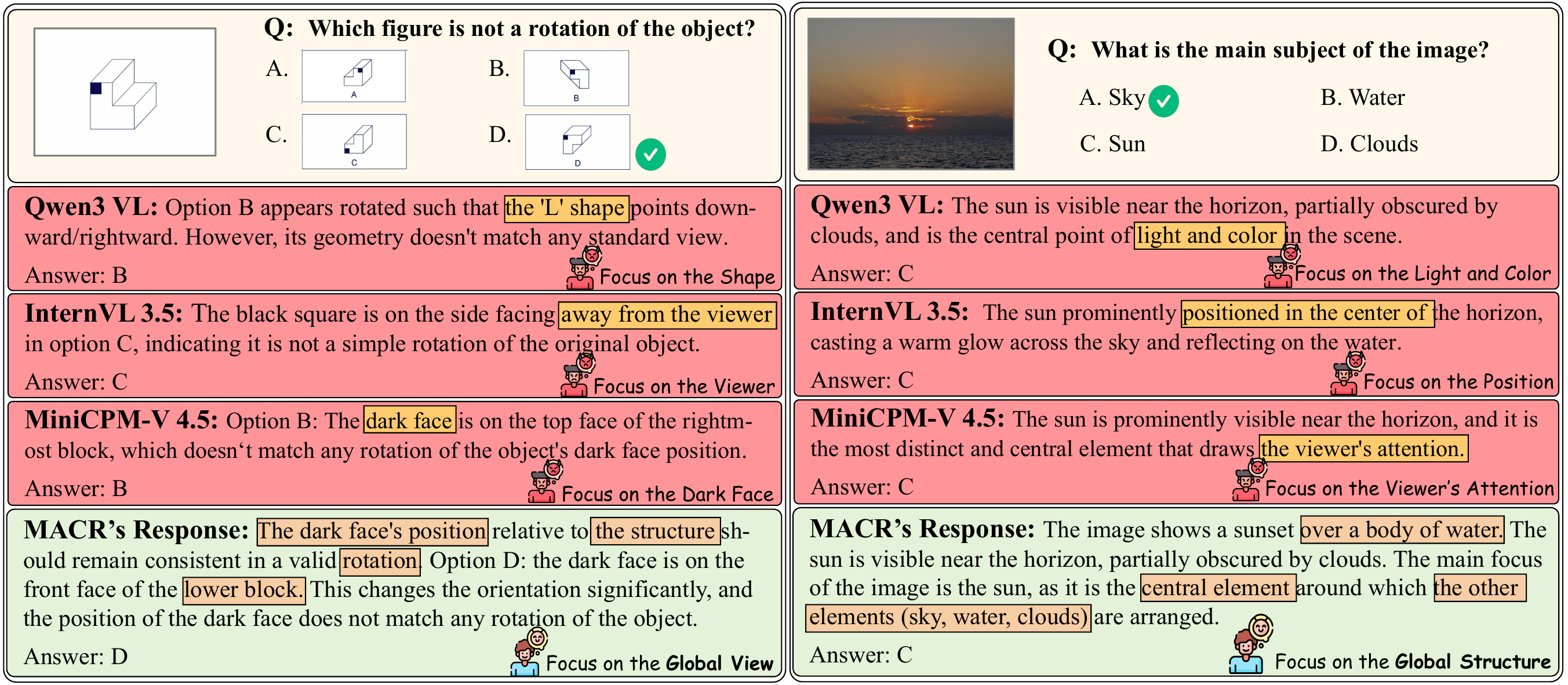}
    \caption{Qualitative comparison between single-models' \texttt{CoT} and \method{}.}
    \label{fig:case}
\end{figure}

\vspace{-5pt}
We compare \method{} reasoning trajectories against those of single-model CoT. In cases where \texttt{CoT} fails (\cf Fig.~\ref{fig:case}, left), single models anchor on isolated cues (\eg, local textures or lighting attributes), whereas \method{} succeeds by synthesizing these fragmented cues into a more comprehensive global view. This advantage extends to failure cases where both methods produce an incorrect answer (\cf Fig.~\ref{fig:case}, right). Even then, \method{} still attempts a comprehensive integration, whereas \texttt{CoT} focuses on an isolated patch. This indicates that \method{} enhances the generation quality by executing global contextual integration, even when the underlying model pool reaches its capability ceiling.

\vspace{-6pt}
\subsection{Analysis}
\label{sec:ana}
\subsubsection{Capability Routing Analysis}
\label{sec:ana1}
To validate the necessity of bidding, we analyze the cross-model transition matrix (\cf Fig.~\ref{fig:ana}a).
The row-normalized transition probabilities reveal bidirectional relay dynamics. While \textit{MiniCPM-V 4.5} has the highest self-transition rate (59.7\%), it still yields dominance to \textit{InternVL 3.5} or \textit{Qwen3 V}L in 40.3\% of its bids.
Furthermore, bidirectional exchanges between peers are frequent. \textit{InternVL 3.5} transfers to \textit{Qwen3 VL} (22.4\%), while \textit{Qwen3 VL} returns control (26.9\%).
This indicates that no single model maintains absolute dominance. Instead, it demonstrates that they complement each other. By leveraging calibrated local confidence, \method{} matches each bottleneck with its optimal model.
Our ablation findings corroborate this complementarity (\cf \S\ref{sec:ablation}).

\subsubsection{Inference Survival Analysis}
\label{sec:ana2}
To quantify the continuous generation capability of models, we conduct a Kaplan-Meier survival analysis (\cf Fig.~\ref{fig:ana}b).
\textit{MiniCPM-V 4.5} shows the highest early survival, with a number at risk (3,151) roughly twice that of the other two models (1,163 and 1,658). \textit{InternVL 3.5} maintains a plateau during later stages, demonstrating long-horizon retention, whereas \textit{Qwen3-VL} exhibits the poorest long-horizon retention.
This indicates that models excel at different stages of generation. Unlike traditional turn-based strategies that may either disrupt long-horizon reasoning or cause early-survival models to accumulate hallucinations during later stages, \method{} relays generation according to each model's stage-wise retention.
Consistent with these temporal dynamics, employing punctuation-based fixed triggers degrades performance (\cf \S\ref{sec:ablation}).

\subsubsection{Task-Agnostic Bottleneck Location Analysis}
\label{sec:ana3}
To validate that \method{} is task-agnostic, we analyze the distribution of bidding triggers across different tasks (\cf Fig.~\ref{fig:ana}c).
The distributions reveal contrasting relay patterns across datasets. While logic-heavy tasks (LogicVista, MathVista) exhibit dense relays in the early stage, visual-intensive tasks (MMStar, HallusionBench) concentrate their relays in later stages. Furthermore, across all tasks, bidding frequency consistently stabilizes during the final stage as models converge on final answers. \method{} localizes these task-agnostic bottlenecks, adapting to the unique reasoning paradigms of different domains. If \method{} adopts task-blind periodic relays, accuracy degrades (\cf \S\ref{sec:ablation}).

\subsubsection{Ablation Study}
\label{sec:ablation}
\newcommand{\qwen}{%
  \adjustbox{valign=c}{\includegraphics[height=2.2ex]{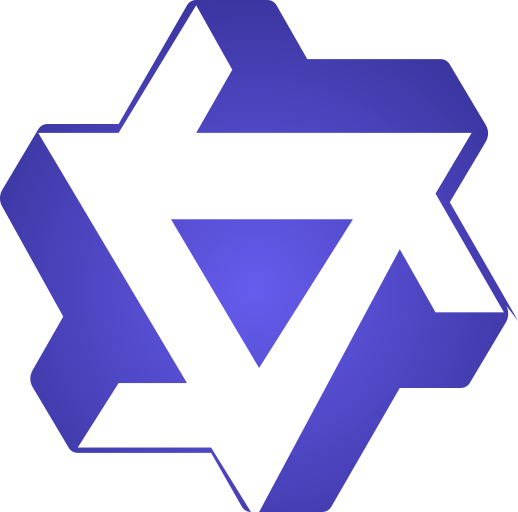}}%
}
\newcommand{\minicpm}{%
  \adjustbox{valign=c}{\includegraphics[height=2.2ex]{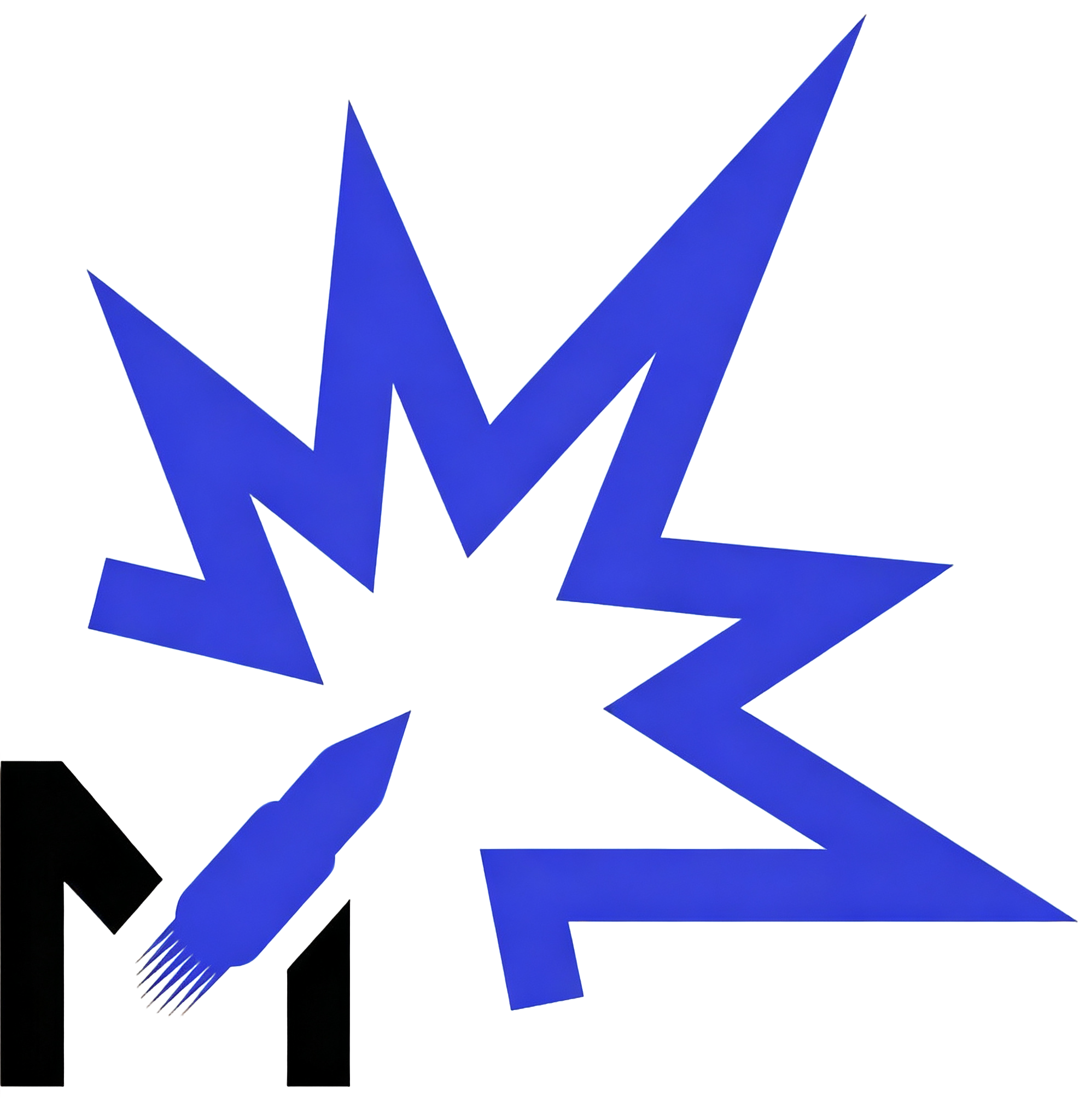}}%
}
\newcommand{\intern}{%
  \adjustbox{valign=c}{\includegraphics[height=2.5ex]{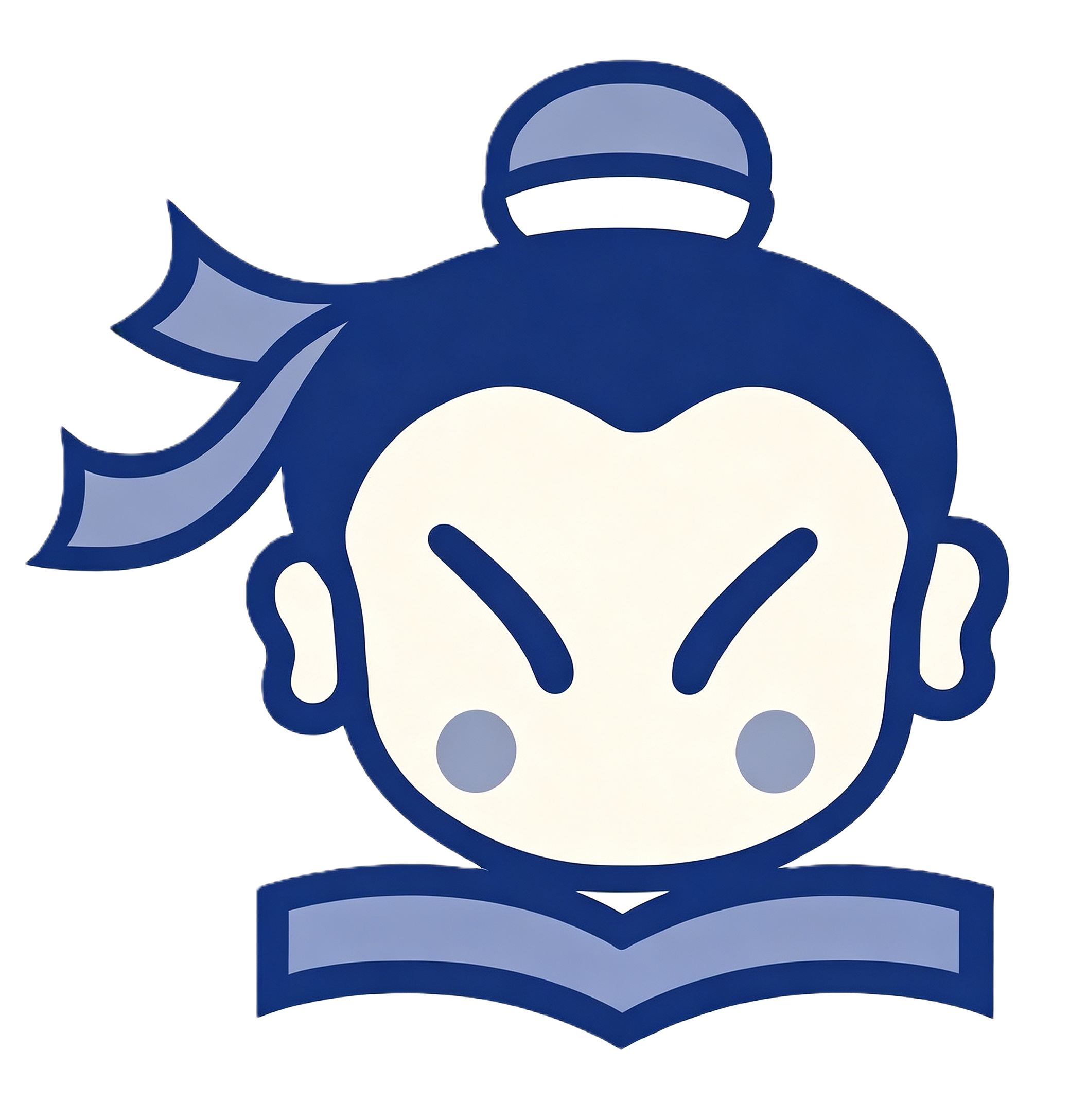}}%
}

\definecolor{lightcyan}{rgb}{.930, .980, .980}
\begin{wraptable}[9]{r}{0.35\textwidth}
\vspace{-15pt}
\footnotesize
\centering
\begin{tabular}{lcc}
\toprule
\textbf{Variant} & \textbf{Acc. (\%)} & \textbf{Tokens} \\
\midrule
\texttt{FI}\hspace{2pt}\minicpm & 71.60 & 374 \\
\texttt{FI}\hspace{1pt}\intern & 71.20 & 322 \\
\texttt{FI}\hspace{2pt}\qwen & 70.40 & 391 \\
\texttt{FS} & 69.00 & 366 \\
\texttt{FT} & 69.80 & 427 \\
\midrule
\rowcolor{lightcyan}
\textbf{\method{}} & \textbf{72.40} & \textbf{312} \\
\bottomrule
\end{tabular}
\caption{Ablation study of \method{}.}
\label{tab:ablation}
\end{wraptable}
To verify the necessity of \method{}'s initiation, capability matching and bottleneck localization, we ablate the respective modules on 500 randomly sampled instances from MMStar.

\textbf{Initiation.} The initial leading model is fixed to each of the three models in turn while keeping the rest of the pipeline unchanged (Fixed Initiation, \texttt{FI}). \texttt{FI} achieves 71.6\%, 71.2\%, and 70.4\% for \textit{MiniCPM-V 4.5}, \textit{InternVL 3.5}, and \textit{Qwen3 VL}, respectively, compared with \method{}'s 72.4\%. This shows that confidence-guided initialization outperforms any fixed pattern.

\textbf{Capability Matching.} The selection mechanism (\cf Eq.~\ref{eq6}) is replaced with a fixed routing order (\textit{MiniCPM-V 4.5} $\rightarrow$ \textit{InternVL 3.5} $\rightarrow$ \textit{Qwen3 VL}) (Fixed Selection, \texttt{FS}). \texttt{FS} degrades accuracy from 72.4\% to 69.0\%. Aligning with the bidirectional exchanges observed in capability routing analysis (\cf \S\ref{sec:ana1}), fixed routing fails to exploit the models' complementarity.

\textbf{Bottleneck Localization.} Confidence-guided bidding (\cf Eq.~\ref{eq5}) is replaced with fixed bidding at every sentence boundary (Fixed Trigger, \texttt{FT}). \texttt{FT} degrades accuracy from 72.4\% to 69.8\%. It is consistent with our findings in inference survival (\cf \S\ref{sec:ana2}) and bottleneck location analysis (\cf \S\ref{sec:ana3}): \texttt{FT} truncates the reasoning chains of long-horizon models while accumulating hallucinations from short-horizon ones, and it cannot adapt to task-specific characteristics.
\newpage

\subsubsection{Analysis on Model Pool Heterogeneity}
\newcommand{\qwenn}{%
  \adjustbox{valign=c}{\includegraphics[height=2.7ex]{figs/qwen.png}}%
}
\newcommand{\minicpmm}{%
  \adjustbox{valign=c}{\includegraphics[height=2.7ex]{figs/minicpm.png}}%
}
\newcommand{\internn}{%
  \adjustbox{valign=c}{\includegraphics[height=3.1ex]{figs/intern.png}}%
}
\setlength{\columnsep}{0.1pt}

\begin{wraptable}[8]{r}{0.46\textwidth}
\footnotesize
\centering
\vspace{-12pt}
\begin{tabular}{lccc}
\toprule
\textbf{Method} & \textbf{\minicpmm} & \textbf{\internn} & \textbf{\qwenn} \\
\midrule
\texttt{RPH} & 68.00 & 70.20 & 64.80 \\
\texttt{TDH} & 66.60 & 68.80 & 63.60 \\
\midrule
\rowcolor{lightcyan}
\textbf{\method{}} & \multicolumn{3}{c}{\textbf{72.40}} \\
\bottomrule
\end{tabular}
\captionsetup{width=0.4\textwidth}
\caption{Analysis on model heterogeneity. All results are reported in Acc. (\%).}
\label{tab:model-pool-heterogeneity}
\end{wraptable}
To validate the necessity of diversity within the model pool, we restrict \method{}'s model pool to a single homogeneous model (\textit{MiniCPM-V 4.5}, \textit{InternVL 3.5}, or \textit{Qwen3 VL}) and evaluate two homogeneous multi-agent collaboration pattern variants on 500 randomly sampled instances from MMStar (\cf Table~\ref{tab:model-pool-heterogeneity}).

\textbf{Role-Playing Pattern} (\texttt{RPH}) prompts a single model with three distinct personas: a Proponent proposing the initial reasoning and answer, an Opponent formulating rebuttals and alternative perspectives, and a Mediator resolving conflicts~\citep{lu2026dialectic}. \texttt{RPH} achieves 68.0\%, 70.2\%, and 64.8\% for \textit{MiniCPM-V 4.5}, \textit{InternVL 3.5}, and \textit{Qwen3 VL}, respectively, compared with \method{}'s 72.4\%. This shows that simulated persona diversity fails to match true architectural complementarity.

\textbf{Temperature-Decoded Pattern} (\texttt{TDH}) simulates intra-model diversity by elevating the sampling temperature during the bidding phase, generating three distinct reasoning paths. The trajectory that maximizes local confidence is then adopted and concatenated to the context. \texttt{TDH} yields 66.6\%, 68.8\%, and 63.6\% for \textit{MiniCPM-V 4.5}, \textit{InternVL 3.5}, and \textit{Qwen3 VL}, underperforming \method{}. This shows that sampling strategies cannot substitute for complementary heterogeneity.

The results indicate homogeneity leads to performance degradation. Consistent with recent findings~\citep{bertalanivc2026cost}, homogeneous pools lack intrinsic diversity and thus cannot produce cognitive diversity. Conformity tendency may drive them to converge on a flawed consensus.

\subsubsection{Lexical Profiling of Reasoning Bottlenecks}
\begin{figure}[t]
    \centering
    \includegraphics[width=0.99\linewidth]{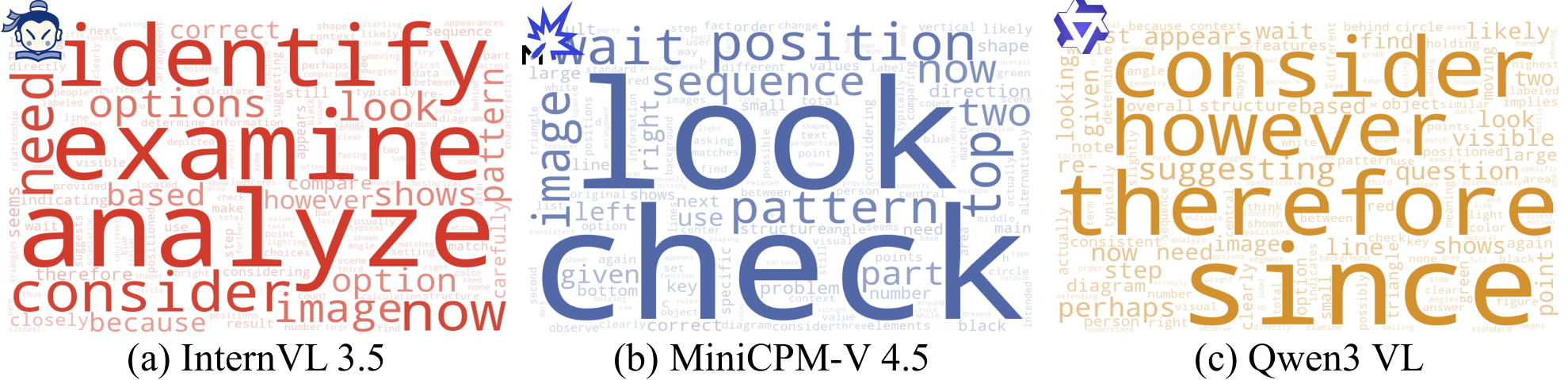}
    \caption{Lexical distribution of three models.}
    \label{fig:words}
    \vspace{-5pt}
\end{figure}

To semantically characterize models' reasoning bottlenecks, we analyze the lexical distribution of tokens generated immediately preceding a bidding trigger (cf. Fig.~\ref{fig:words}). The resulting distributions show distinct lexical patterns across models. \textit{InternVL} predominantly falters around analytical evidence extraction (e.g., \textit{analyze, examine, identify}), \textit{MiniCPM-V} around spatial localization and attribute binding (e.g., \textit{position, top, left, large}), and \textit{Qwen3-VL} around causal deduction (e.g., \textit{since, therefore, suggesting}). Despite these differences, all models share a universal hesitation signature, frequently outputting stall tokens (e.g., \textit{wait, look, check}). This lexical heterogeneity aligns with the complementarity we observed (cf. \S\ref{sec:ana1}), suggesting that different models tend to hesitate at different stages and motivating \method{}'s design of matching each bottleneck with its optimal model.
\vspace{-15pt}
\section{Related Work}
\noindent\textbf{Multi-Agent Debate (MAD)} improves performance by leveraging interactions among LLMs. Existing MAD methods are broadly categorized into two streams.
The first category is \textit{Collaboration Workflow Optimization}~\citep{hong2024metagpt, maiga-etal-2025-error, li2024improving, pham2024let, zhang-etal-2024-self-contrast, li2023prd, zhang2025debate4math}. These methods structure multi-agent collaboration to resolve tasks, \eg, orchestrating role-based coordination pipelines~\citep{hong2024metagpt, maiga-etal-2025-error}, communicating via continuous embeddings~\citep{pham2024let}, and optimizing communication topologies or multi-perspective reward frameworks~\citep{li2024improving, zhang-etal-2024-self-contrast, li2023prd, zhang2025debate4math}.
The second category is Failure Diagnosis and Intervention~\citep{wynn2025talk, yao2025peacemakertroublemakersycophancyshapes, cemri2026multi, choi2026debate, yang2025revisiting}. These methods diagnose and mitigate collaborative limitations, \eg, quantifying sycophancy and social conformity to prevent premature convergence~\citep{wynn2025talk, yao2025peacemakertroublemakersycophancyshapes}, establishing taxonomies for system misalignments~\citep{cemri2026multi}, and delineating theoretical bounds of debate under test-time scaling~\citep{choi2026debate, yang2025revisiting}. Despite their impressive performance, these methods are all built around the topology of \textit{group thinking}, where models produce complete responses that are aggregated after generation.

Unlike methods relying on rigid debate turns, \method{} monitors confidence in real time and relays generation rights to the most confident candidate when a model struggles.

\textbf{Uncertainty and Confidence in LLMs} provide signals for evaluating confidence and guiding reasoning. These approaches are broadly divided into two main streams.
The first category is Uncertainty Quantification and Decomposition~\citep{wang2025measuring, jayasekera2026variational, feng-etal-2025-rethinking, amayuelas2024knowledge, liu2026let}. They decompose and estimate uncertainty to evaluate confidence or trigger behaviors, \, establishing bounds or benchmarks for uncertainties~\citep{jayasekera2026variational, wang2025measuring}, inducing uncertainty awareness through query or tuning~\citep{feng-etal-2025-rethinking, amayuelas2024knowledge}, and extracting token-level logit margins to dynamically trigger interleaved multimodal reasoning~\citep{liu2026let}.
The second category is Confidence-Guided Collaboration and Reinforcement~\citep{zhu2026demystifying, ai2025beyond, yang2026understanding, pappu2026multi, li2026m3mad, choi2026identity, li2026dynadebate, liu2026llm, park2025maporl, zhao2026stronger}. They manage uncertainty within multi-agent frameworks, \eg, modulating belief updates through confidence scores~\citep{zhu2026demystifying, ai2025beyond, yang2026understanding}, mitigating behavioral uncertainties like hallucination, compromise, or bias during debates~\citep{pappu2026multi, li2026m3mad, choi2026identity, li2026dynadebate}, and optimizing collaborative policies via translating confidence into reward signals~\citep{liu2026llm, park2025maporl, zhao2026stronger}.
Despite quantifying uncertainties and optimizing collaborative policies, these methods evaluate sequence-level outputs and overlook the comparability of confidence across heterogeneous models.

Unlike these methods, we first identify the incomparability of confidence across heterogeneous models and provide a calibration method for it. Building on this, \method{} calibrates local confidence at the window level rather than sequence level, monitoring mid-generation uncertainty and executing cross-model relays the moment a local reasoning bottleneck is detected.

\vspace{-1pt}
\section{Conclusion}
\vspace{-4pt}
\label{sec:conclusion}
We introduce \method{}, transitioning multi-agent collaboration from \textit{group thinking} mode to \textit{relay thinking} mode. Addressing incomparable confidence across heterogeneous models, \method{} calibrates local confidence, enabling real-time cross-model bidding. Monitoring the leading model, \method{} interrupts generation upon detecting reasoning bottlenecks, transferring generation rights to the most confident model. Experiments across four multi-modal benchmarks demonstrate that \method{} outperforms all existing single-model baselines in accuracy, while achieving results comparable to existing multi-agent baselines with 6.0$\times$ less token overhead. Further analyses verify that this relay mechanism integrates isolated cues into comprehensive global reasoning chains and exploits model complementarity, with a task-agnostic selection mechanism.
\newpage


\bibliography{iclr2027_conference}
\bibliographystyle{iclr2027_conference}

\end{document}